\documentclass{ecai}  % use option [doubleblind] for double blind submission and hiding the authors section

\usepackage{graphicx}
\usepackage{latexsym}
\usepackage{hyperref}
\usepackage{caption}  % in the preamble
\usepackage{bm}                 % For bold in results tables
\usepackage{subcaption}         % For multiple sub plots
\usepackage{verbatim}
\usepackage{amsmath}            % For writing mathematical formulas

\begin{document}

\begin{frontmatter}

\title{FuelCast: Benchmarking Tabular and Temporal
Models for Ship Fuel Consumption}

\author[A]{\fnms{Justus}~\snm{Viga}}
\author[A]{\fnms{Penelope}~\snm{Mueck}}
\author[B]{\fnms{Alexander}~\snm{Löser}}
\author[C]{\fnms{Torben}~\snm{Weis}}

\address[A]{KROHNE Messtechnik GmbH, Duisburg, Germany}
\address[B]{Berliner Hochschule für Technik, Berlin, Germany}
\address[C]{University of Duisburg-Essen, Duisburg, Germany}

\begin{abstract}
In the shipping industry, fuel consumption and emissions are critical factors due to their significant impact on economic efficiency and environmental sustainability. Accurate prediction of ship fuel consumption is essential for further optimization of maritime operations. However, heterogeneous methodologies and limited high-quality datasets hinder direct comparison of modeling approaches. This paper makes three key contributions: (1) we introduce and release a new dataset
(\url{https://huggingface.co/datasets/krohnedigital/FuelCast}) comprising operational and environmental data from three ships; (2) we define a standardized benchmark covering tabular regression and time-series regression (3) we investigate the application of in-context learning for ship consumption modeling using the TabPFN foundation model - a first in this domain to our knowledge. Our results demonstrate strong performance across all evaluated models, supporting the feasibility of onboard, data-driven fuel prediction. Models incorporating environmental conditions consistently outperform simple polynomial baselines relying solely on vessel speed. TabPFN slightly outperforms other techniques, highlighting the potential of foundation models with in-context learning capabilities for tabular prediction. Furthermore, including temporal context improves accuracy.
\end{abstract}

\end{frontmatter}

% Include content from sections/intro.tex

\begin{figure}[ht]
    \centering
    \includegraphics[width=0.47\textwidth]{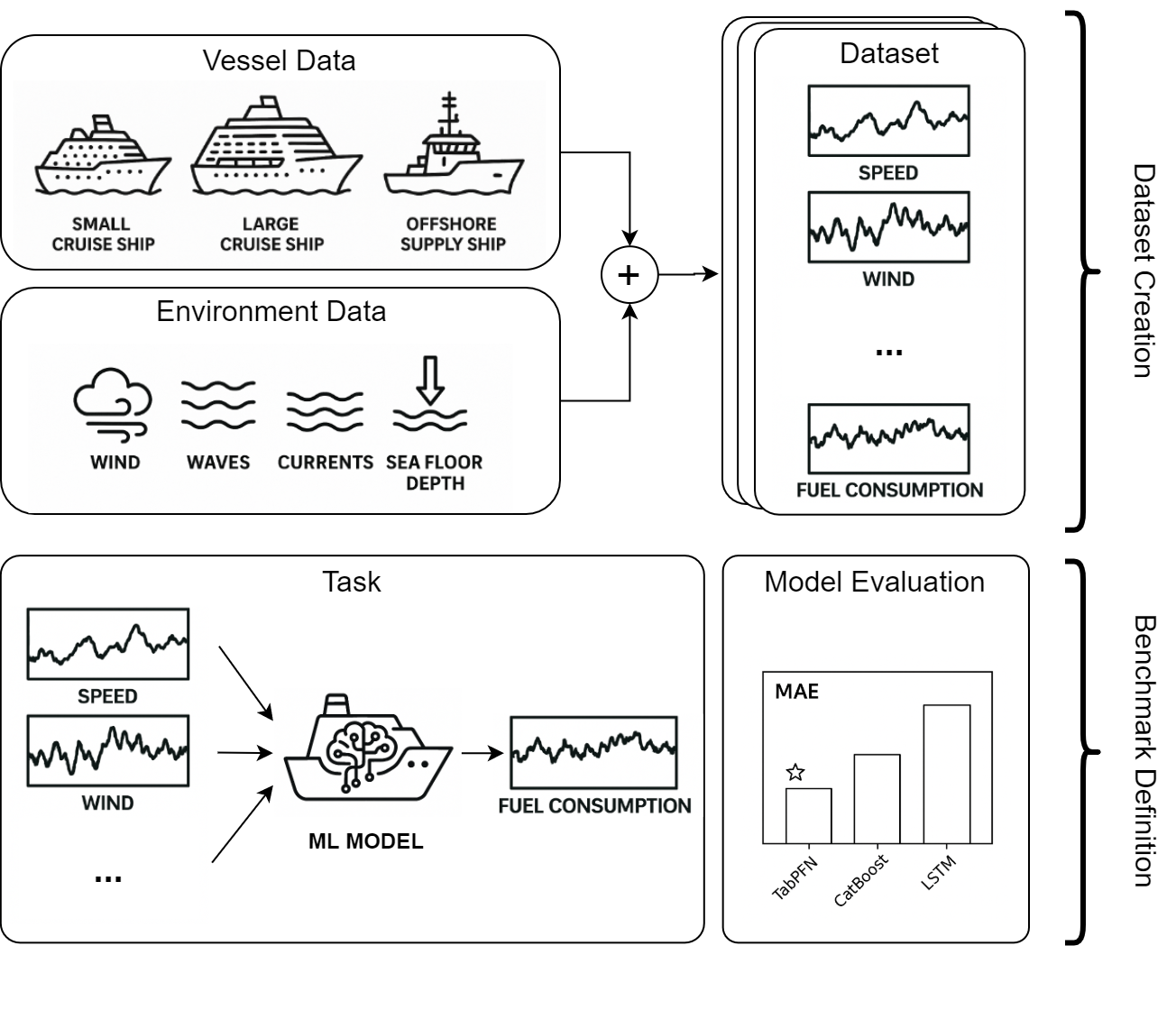}
    \caption{FuelCast benchmark: We create a contextual rich benchmark dataset from operational data of three ships and environmental data. We define timeseries regression tasks for fuel consumption prediction and an evaluation setup. From the models that we apply, with in-context learning model TabPFN for the first time in this domain, TabPFN outperforms other models.}
    \label{fig:fuelcast}
\end{figure}

\section{Introduction}
In the shipping industry, fuel consumption and emissions are key performance indicators with far-reaching economic and ecological consequences. With stricter international regulations and rising climate concerns, reducing emissions has become a strategic imperative. Accurate fuel consumption prediction plays a central role in this context, enabling optimized routing, operational planning, and emissions estimation. However, predictive modeling in maritime settings faces fundamental challenges: data scarcity, high variability, and the lack of standardized benchmarks. 

Approaches to fuel optimization vary widely, encompassing physical models based on domain knowledge, statistical modeling, and data-driven machine learning methods, each with different priorities, assumptions, and data needs.  This diversity complicates comparisons and limits reproducibility. Standardized benchmarking for temporal data is crucial to consistently evaluate and compare models.

Although some publicly available datasets exist \cite{conference:petersen2011}, \cite{pekka_siltanen_2019_3563390} they are typically limited to a single vessel type, span only short time periods, lack detailed context data or simulate behavior for limited settings \cite{DBLP:data/10/BaierS22a}. As a result, they fall short of supporting robust model evaluation for long-term, generalizable fuel consumption prediction.

Ideally, benchmark datasets for fuel consumption prediction in maritime transport would span multiple years, cover a diverse fleet of vessels, and include rich contextual information such as weather conditions, sea states, engine parameters, operational modes, and route metadata. They would be accurately labeled, time-synchronized, and representative of real-world variability across seasons, vessel types, and operational patterns. Furthermore, such datasets would provide annotations for known anomalies or regime changes.% and be openly accessible to the research community.

However, creating such datasets is extremely challenging. Maritime operational data is often fragmented across stakeholders, stored in proprietary formats, and subject to strict confidentiality constraints. Ensuring data quality through cleaning, synchronization, and contextualization requires significant domain expertise and infrastructure.

Additionally, legal and commercial concerns frequently prevent open publication. 
As a result, comprehensive, high-quality, and openly available temporal datasets in the maritime sector remain the exception.

To address this limitation, we introduce a new benchmark dataset and evaluation protocol that aims to support the development and assessment of time series models, see Figure \ref{fig:fuelcast}.

Our primary audience is the machine learning research community, with a focus on temporal modeling in realistic operational settings and enabling cross-vessel comparisons. The dataset spans multiple vessel types and includes rich contextual features such as weather data, making it one of the few openly available resources of its kind.

We evaluate three representative model families with complementary strengths: CatBoost \cite{dorogush2018catboostgradientboostingcategorical}, a gradient boosting method well-suited for structured tabular data; LSTM \cite{lstm1997}, a classical deep learning approach for sequence data; and TabPFN \cite{hollmann2023tabpfntransformersolvessmall}, a pretrained probabilistic transformer designed to generalize across tabular tasks without fine-tuning.  Additionally, we include simple baselines such as polynomial regression and a multilayer perceptron (MLP) \cite{Goodfellow-et-al-2016} for selected tasks to provide reference points for model complexity and data efficiency. This setup allows us to investigate a central question: Is successful fuel consumption modeling primarily dependent on large-scale supervised training, or can pretrained foundation models offer strong performance even with limited data? We deliberately omit standard Transformer architectures, as their typical reliance on large datasets and tuning budgets contrasts with our focus on practical, data-efficient approaches.

\textbf{In this work}, we contribute 

\begin{enumerate}
    \item A new long-term timeseries dataset comprising operational and environmental data from three ships
    \item A structured evaluation protocol across two time series regression tasks, and
    \item Apply in-context learning using the TabPFN foundation model, which, to the best of our knowledge, has not yet been applied in maritime fuel prediction and represents a novel approach to modeling low-data and complex tabular tasks.
\end{enumerate}

Our dataset and tasks are intended to support research in representation learning, forecasting, and explainable modeling for temporal data in energy-intensive domains. By releasing them publicly, we aim to foster exchange between machine learning and maritime communities, and to provide a realistic foundation for developing and benchmarking new models.

The remainder of this paper is structured as follows: Section \ref{related-work} reviews related work. Section \ref{experimental-setup} introduces the vessels, describes the dataset, and outlines the benchmark setup, including tasks, models, and evaluation protocol. Section \ref{results} presents the experimental results. Section\ref{discussion} discusses findings and limitations. Finally, Section \ref{conclusion} concludes the paper.

% Include content from sections/related-work.tex
\section{Related Work} \label{related-work}

\subsection{Data Sets}
Public datasets of operational ships with high-frequency, long-term data are rare. Authors of \cite{conference:petersen2011} provide two months of high-resolution ferry data, including wind sensor measurements. Authors of  \cite{pekka_siltanen_2019_3563390} published one-month datasets for three fishing vessels at one-minute intervals, with detailed ship operational and wind data. While authors of \cite{DBLP:data/10/BaierS22a} provide a simulated dataset for one ship under different weather conditions. To our knowledge, our three datasets are the first to offer long-term, high-resolution operational ship data. Unlike \cite{conference:petersen2011} and \cite{pekka_siltanen_2019_3563390}, we also include comprehensive environmental conditions for ships of two different types. In contrast to \cite{DBLP:data/10/BaierS22a} our datasets cover real-world long term effects.

\subsection{Tasks and Models}
Modeling fuel consumption requires features from multiple sources, including ship sensors, weather, and current data. To capture a contextual representation of the vessel, we explore both manual feature engineering and neural models using mostly raw data. On top of these, we apply regression methods.

\paragraph{Regression} 
Linear regression and similar models \cite{WANG2018817}, \cite{CEPOWSKI2023121315}, ensemble methods \cite{ZHOU2023115509}, SVM \cite{DU2022100072}, \cite{GKEREKOS2019106282} and ANNs \cite{pedersen2009}, \cite{ZHOU2023115509}, \cite{GKEREKOS2019106282} have been widely studied for fuel consumption prediction. We implement a simple third-order polynomial regression model motivated by the admiralty coefficient commonly used in ship performance analysis.

Ridge and Lasso regression \cite{tibshirani96regression} are limited in modeling non-linearities, motivating the use of gradient boosting methods such as CatBoost \cite{dorogush2018catboostgradientboostingcategorical}, which handle outliers well and are less prone to overfitting.

\paragraph{Neural Networks and Time-Aware Models} Some approaches explicitly incorporate time, e.g., by splitting data into non-overlapping windows \cite{pedersen2009}. %, with optimal window sizes found around 3 minutes. 
LSTMs have been applied to predict engine speed and fuel consumption \cite{ZHANG2024107425}, \cite{Lei_2021}. Authors of \cite{10325522} use transformer-based approaches to predict main and auxiliary engine fuel consumption. Authors of \cite{AGAND2023115271} apply LSTMs for multi-step prediction. 

Multilayer Perceptrons (MLPs) can approximate complex non-linear functions but do not model temporal dependencies and often require regularization and large datasets \cite{Goodfellow-et-al-2016}. Time series forecasting traditionally builds on autoregressive models, where future values are predicted based on past observations. These approaches motivate the use of lag features and moving windows in non-sequential models.
To learn temporal dependencies directly, we employ Long Short-Term Memory Networks (LSTMs) \cite{lstm1997}, which capture both short- and long-term patterns through internal memory states and are well-suited for multivariate time series prediction.
Most recently authors of \cite{hoo2025tabularfoundationmodeltabpfn} introduced a foundation model for tabular regression at test-time compute. It enables fast regression without hyperparameter tuning and outperforms classical methods for many datasets \cite{hollmann2023tabpfntransformersolvessmall}. To our best knowledge, we are the first, investigating test-time-compute/in-context learning methods for fuel consumption prediction on ships. 

\subsection{Benchmarks}
To our best knowledge, we provide the first comprehensive benchmark for fuel consumption prediction using ships and their operational context. Few methods \cite{10325522}, \cite{pedersen2012a} use  public datasets \cite{pekka_siltanen_2019_3563390}, \cite{conference:petersen2011}. However, prior work lacks rigorous method comparison.

\paragraph{Fuel Oil Consumption Prediction}
Fuel consumption prediction models in shipping are typically based on physical principles, data-driven methods, or both. They commonly use noon reports, AIS data, onboard sensors, and contextual information like maintenance records and environmental conditions. However, prior work is highly heterogeneous in terms of data quality, sources, and modeling approaches.
For instance, \cite{DU2022100072} combine high-frequency sensor data from two container ships with external weather data from Copernicus Marine and ECMWF, achieving best results using a rich feature set including vessel-relative wind and current information. While our data construction is similar, we construct and publish higher-resolution datasets and use them primarily for rigorous method comparison rather than single-model performance.
Other studies vary in scale and scope: Authors of \cite{WANG2018817} use operational and mechanical data from many voyages; authors of \cite{pedersen2009} and \cite{pedersen2012a} rely on data from a single tanker or ferry; and some, like authors of \cite{CEPOWSKI2023121315} and \cite{CORADDU2017351}, incorporate additional factors such as hull cleaning. Models are typically applied either to individual vessels or aggregated across fleets. Target variables also differ widely—from hourly fuel consumption to per-voyage or engine-specific predictions \cite{YAN2021102489}, \cite{GKEREKOS2019106282}, \cite{10325522}.
While some works propose evaluation frameworks \cite{GKEREKOS2019106282}, \cite{BASSAM2022110449}, consistent comparison remains difficult due to the lack of public datasets, varying experimental setups, and inconsistent prediction targets.

In contrast, our benchmark addresses these limitations by providing publicly available, long-term high-resolution datasets with a unified comparison framework. We define fixed tasks and targets relevant to real-world shipping operations and apply a diverse set of machine learning models, including, uniquely in this domain, test-time compute methods. Our datasets span multiple vessel types, enabling systematic evaluation of model generalization across ships.

% Include content from sections/experimental-setup.tex
\section{The FuelCast Benchmark} \label{experimental-setup}

This section provides an overview of our benchmark setup. First, we introduce the three vessels and explain the construction of our datasets.\textbf{ Note that ship names are completely anonymized.}  Next, we derive the tasks we consider and briefly introduce the machine learning models we apply in our experiments. Finally, we describe our benchmark setup and the experiments we conduct in more detail for each task and model.

\subsection{Ships}
We provide data for the following three ships, see also Table \ref{tab:basic_dataset}: A small cruise passenger ship CPS Triton, a large cruise passenger ship CPS Poseidon and an offshore supply ship OSS Ceto. All ships names follow the naming scheme abbreviation of the type, e.g. CPS for cruise passenger ship, whitespace and individual name.

\begin{table}[ht]
    \captionsetup{justification=centering}  % center the caption
    \caption{Overview of the ships and corresponding datasets. We provide data from three ships with different sizes and types to compare the domain generalization performance of the models.}
    \label{tab:basic_dataset}
    \centering
    \resizebox{0.5\textwidth}{!}{
        \begin{tabular}{l l l l l l}
            \hline
            \\[-6pt]
            Ship & Type & \shortstack{Gross\\Tonnage} & \# Samples & \shortstack{Missing\\Values} \\
            \hline
            \\[-6pt]
            CPS Triton & Cruise Passenger Ship & 11,000 & 25,351 & 0.04 $\%$\\
            CPS Poseidon & Cruise Passenger Ship & 70,000 & 105,422 & 3.2 $\%$ \\
            OSS Ceto & Offshore Supply Ship & 24,000 & 43,213 & 0.96 $\%$\\
            \hline
            \\[-6pt]
        \end{tabular}
    }
\end{table}

\paragraph{The CPS Triton} is a cruise passenger ship that operates on a fixed several-day route. She is powered by two diesel engines with a straight shaft to the propellers. For this ship we provide three months of data.

\paragraph{The CPS Poseidon} is a cruise passenger ship with an area of operation that is more diverse and has a strong seasonality. She has a diesel-electric propulsion system with five generator engines and a straight shaft to the propellers from electric motors. We provide 12 months of data.

\paragraph{The OSS Ceto} is a ship that supports deep sea operations. Her operation profile is very different compared to the cruise passenger ships with many short maneuvering sections. She has intervals of dynamic positioning where the engines are active to work against ocean currents and winds to keep her at a fixed position. She has thruster pods with electric motors that are powered by diesel generators. We provide 6 months of data.

\subsection{Datasets}
For the construction of our datasets, one for each vessel, we consider operational ship data and environmental data. In the following, we also use the vessels name to refer to the corresponding dataset. We integrate the two data sources by time and position. Table \ref{table:dataset-features} shows the data used in our experiments.

\paragraph{Vessel-Specific Data} By ship operational data we refer to data produced by all processes directly connected to the ship itself. The data is collected by onboard sensors with a sample rate of five minutes. It contains information about time and position, speed, the ship's heading or bearing and detailed consumption data per consumer. Each ship has multiple consumers that contibute to the total momentary fuel consumption. These can be engines for propulsion and power generation or other consumer like boilers and incinerators. For the measurement of consumption we use accurate KROHNE Coriolis mass-flow meters. For each engine we measure inlet and outlet and combine these using the difference to calculate exact consumption.

\paragraph{Sea- and Weather Data} The sea and weather condition data includes sea temperature, depth and current as well as details on wind, waves and external temperature. We use satellite data provided by Copernicus Marine\footnote{https://marine.copernicus.eu} (sea floor depth) and Open-Meteo \cite{Zippenfenig_Open-Meteo}. The data points are provided on a coarse coordinate grid with a sample rate of one hour.

\begin{table}[ht]
\captionsetup{justification=centering}  % center the caption
\caption{Overview of variables from the datasets that we use in our experiments including the name, a short description, source and the unit. Momentary fuel and speed over ground are integrated by position and time with the sea floor depth information from Copernicus Marine and the historical weather data from Open-Meteo. For the experiments we transform directional features relative to the ship.}
\label{table:dataset-features}
\centering
\resizebox{0.5\textwidth}{!}{
\begin{tabular}{l p{2.5cm} p{1.5cm} l} 
\hline
\\[-6pt]
Variable & Description & Source & Unit \\
\hline
\\[-6pt]
 Total.MomentaryFuel & Total momentary fuel consumption of all consumers on the vessel. & Flowmeter & $kg/s$ \\
 SpeedOverGround & Speed over ground of the vessel. & GPS & $m/s$ \\ 
 SeaFloorDepth & Sea floor depth below sea level (bathymetry). & Copernicus Marine & $m$ \\
 WindDirection10M & Wind direction at 10 meters above ground. & Open-Meteo & $^\circ$ \\
 WindSpeed10M & Wind speed at 10 meters above ground. & Open-Meteo & $m/s$ \\
 OceanCurrentDirection & Ocean current direction considering all components. & Open-Meteo & $^\circ$ \\
 OceanCurrentVelocity & Ocean current velocity considering all components. & Open-Meteo & $m/s$ \\
 WaveDirection & Mean direction of significant waves. & Open-Meteo & $^\circ$ \\
 WaveHeight & Significant mean wave height. & Open-Meteo & $m$ \\
 WavePeriod & Period between significant waves. & Open-Meteo & $s$ \\
 Temperature2M & Air temperature 2 meters above ground. & Open-Meteo & $m$\\
\hline
\\[-6pt]
\end{tabular}
}
\end{table}

\paragraph{Integrating Data Sets} We integrate ship operational data and weather and sea condition data by time and position. Since the time and position resolution for the two data sources differs, we first linearly interpolate the weather and sea condition data to get data on a finer time and coordinate grid. Once we have 5-minute resolution weather and sea condition data, we integrate the data from both sources by time and position.

\paragraph{General Preprocessing} Due to measurement errors and missing weather data points close to the shore missing values can occur. For the OSS Ceto and the CPS Triton, we observed a percentage of missing values between 0.06$\%$ and 0.2$\%$ and for CPS Poseidon between 0.2$\%$ and 2$\%$. Most missing values occured for bearing while we do not observe missing values for the target value.
We apply column-wise mean imputation to handle missing data. All features are normalized using a standard scaler. Directional features (wind, wave, and ocean current directions) are transformed into the vessel’s local coordinate system by subtracting the vessel's bearing. 

\subsection{Tasks}
We propose two tasks: The goal for all tasks is to predict the total fuel consumption of one vessel. We consider a tabular regression task and a time-based regression task.
\paragraph{Task 1: Tabular Regression}
For this task we consider the classical supervised learning regression task. Given a set of inputs $X$, predict a real-valued variable $y$. In the given case $X$ is a subset of ship operational data and environmental condition data and $y$ the total fuel consumption.

This scenario represents a pointwise prediction based solely on features observed at a single time step, without incorporating temporal context. It abstracts the vessel’s dynamic behavior under the assumption that the dominant influencing factors are stationary or sufficiently reflected in the instantaneous measurements. This task is relevant for assessing how operational parameters such as engine load or vessel speed affect fuel consumption and can support decisions in steady cruising conditions.

\paragraph{Task 2: Timeseries Regression}
Predict $y_t$ from current and past inputs $x_t, ..., x_{t-k}$ with window T: 
\begin{equation}
    y_t = f(x_t, x_{t-1}, .., x_{t-k}) + \epsilon_t
\end{equation}
where $f(.)$ is a function that describes the model, $\epsilon_t$ an error term and $1 \le k \le T$. For example, $y_t$ is the fuel consumption and $x_{t-k}$ describes the collection of speed and wind $k$ observations before $t$.%\\

This setting relaxes the stationarity assumption of task 1 by incorporating recent temporal context from the ship and its surrounding environment. Including historical input patterns enables the model to learn dynamic behavior, such as acceleration phases or maneuvering. It is particularly valuable for analyzing fuel consumption across complete voyages and for understanding the influence of short-term transitions on vessel performance.

\subsection{Benchmark Setup}

\paragraph{Feature Selection}
We select the input features based on physical knowledge and manual feature importance analysis.
The input features are SpeedOverGround, SeaFloorDepth, Temperature2M, OceanCurrentVelocity, OceanCurrentDirection (transformed to local vessel coordinates), WindSpeed10M, WindDirection10M (transformed to local vessel coordinates), WaveHeight, WavePeriod, WaveDirection (transformed to local vessel coordinates). 
As target variable, we use the vessel’s total momentary fuel consumption. 

\paragraph{Data Split}
We segment the dataset into 5 disjoint temporal intervals of equal length, assuming each interval represents an independent realization of vessel behavior (i.i.d. assumption across intervals). These intervals form the basis of a 5-fold cross-validation scheme. We construct batches from each interval that are derived from the scenario of the task. Task 1 requires no special treatment. For task 2, we set the context size to $T=11$. Together with the observation at time t, we get 1 hour windows. We take a sliding window approach with a stride of 1 to create the final batches. For each cross-validation run, we take one of the intervals as a test set and combine the batches of the remaining intervals to form the training set. This procedure ensures that the temporal structure within intervals is preserved, while avoiding leakage across folds. Finally, the samples within each training and test set are shuffled to break the temporal relation between consecutive observations and prevent the models from "cheating".

\paragraph{Metrics} We evaluate the performance of our models using the mean average error (MAE) to calculate the error and R² to calculate how well the variance within the data has been captured by the model.

\subsection{Models}

We evaluate a range of models across the two benchmark tasks to compare simple baselines, classical machine learning methods, and modern and foundation models.

For task 1, we use a third-order polynomial regression model as a speed-based baseline, motivated by the admiralty coefficient commonly used in ship performance analysis. Additionally, we include CatBoost which is well suited for structured data and can capture non-linear relationships effectively. MLPs are tested as general-purpose function approximators, and TabPFN, a transformer-based foundation model designed for tabular data, is included to explore the feasibility of in-context learning for ship fuel prediction.

For task 2, we evaluate CatBoost, LSTM, and TabPFN. CatBoost is extended with lag-based features such as the mean vessel acceleration and the rate of change in sea floor depth to represent temporal dependencies in a static feature format. LSTM is employed as a standard architecture for sequential modeling, allowing the model to directly capture short-term temporal dynamics from sequences of past observations. TabPFN, originally designed for tabular data, is evaluated on time series by using the same lagged feature format as CatBoost, allowing us to assess its ability to model temporal dynamics without recurrence or attention.

All models are configured consistently for fair comparison. CatBoost uses internal feature importance for feature selection and a 3-fold grid search for tuning major hyperparameters. MLP is designed with 20 layers of 32 neurons each, using GELU activations and a final linear output layer. LSTM includes a single recurrent layer with 128 hidden units. Both MLP and LSTM are trained using the mean squared error (MSE) loss and the Adam optimizer with a learning rate of $10^{-3}$. Training includes early stopping based on validation loss, with 20\% of the training set reserved for validation. For MLP and LSTM, directional input features (wind, wave, ocean current) are decomposed into sine and cosine components. TabPFN is applied without architectural changes or additional training. Due to limited computational resources and to evaluate the data efficiency of TabPFN, we restrict the training context to a randomly selected subset of 500 samples (TabPFN (500)) or 1000 samples (TabPFN (1000)).

% Include content from sections/results.tex
\section{Results} \label{results}
In this section, we present empirical results addressing the key questions motivating our benchmark design. Specifically, we investigate (i) the role of temporal context for accurate fuel consumption prediction, (ii) the potential of in-context learning with foundation models—in particular, TabPFN with little data, as a novel approach in maritime operational settings.
(iii) the performance of different modeling paradigms across tabular and time-series tasks, and (iv) the influence of environmental conditions, particularly weather, on fuel consumption. 

\paragraph{Average Results Over All Vessels}
When averaging performance across all vessels, see last column of Tables \ref{tab:results_task_1} and \ref{tab:results_task_2}, clear model hierarchies emerged. TabPFN with 1000 training samples consistently achieved the lowest MAE across all three tasks, followed by its 500-sample variant. In tasks 1 and 2, CatBoost generally ranked third behind the two TabPFN variants. For R² scores, MLP led in task 1, with CatBoost and TabPFN close behind. In task 2, CatBoost showed the highest R², while TabPFN 500 had the weakest performance.

\paragraph{TabPFN Outperforms Across Vessels and Tasks} A high-level analysis of the benchmark results observed in Figure \ref{fig:results_task_1} reveals several consistent trends across vessels, tasks, and model types. Overall, TabPFN achieves the most robust performance, outperforming other models in most tasks and across all three vessels. Its advantage is particularly clear when trained on larger datasets (e.g., with 1000 samples), where it consistently achieves the lowest MAE. This suggests that TabPFN generalizes well across different vessel types and operational conditions, making it a strong candidate for maritime fuel consumption prediction tasks.

\paragraph{Temporal Information Boosts Model Performance} The addition of temporal information through time-based features leads to further improvements in prediction accuracy across most tasks, see trends from left to right plots for the two metrics in Figure \ref{fig:results_task_1}. This underlines the importance of capturing dynamic patterns in vessel behavior and external conditions over time. 

\paragraph{Environmental Features Improve Prediction Accuracy} The speed-based polynomial baseline consistently underperforms. It fails to capture key external drivers of fuel consumption. In contrast to models that include environmental variables, we observe much higher MAE and lower R² in Table \ref{tab:results_task_1}.

\paragraph{Vessel-Specific Variability in Model Performance} Model performance varied notably across the evaluated vessels, with OSS Ceto exhibiting the highest variance in tasks 1 and 2. This variability likely reflects more complex operational profiles, data inconsistencies, or noisier fuel consumption patterns compared to the cruise passenger ships. CPS Triton consistently shows the lowest prediction errors due to more stable operations resulting from the fixed route. Conversely, the CPS Poseidon has higher absolute errors despite strong R² scores, indicating systematic but more complex consumption behavior.

We now investigate the the performance of different models per task.

\paragraph{Task 1: Regression} In task 1 we observe in Table \ref{tab:results_task_1} in the two bottom rows that TabPFN consistently achieved the lowest MAE across all vessels. For CPS Poseidon, it reached 0.061 (1000 samples), ahead of MLP (0.066), CatBoost (0.068), and the polynomial baseline (0.086). R² values among the top models ranged from 0.93 to 0.94. On the CPS Triton, TabPFN again led (MAE = 0.019, R² = 0.859), with CatBoost slightly ahead of MLP, unlike in the large cruise ship case. For OSS Ceto, TabPFN achieved the best MAE (0.042), though MLP had the highest R² (0.547). Variance in R² across folds was high, especially for this vessel, but CatBoost and MLP showed the most stable results. Overall, TabPFN proved most reliable in minimizing absolute error, while relative model performance varied by vessel and metric.

\begin{table*}[ht]
\captionsetup{justification=centering}
\caption{Results of task 1 (tabular regression): Comparison of the evaluated models in terms of MAE and R² metrics across all three vessel datasets. The values are averaged over the 5 cross-validation folds. The standard deviation is given in brackets. TabPFN slightly outperforms the other models in most cases.}
\label{tab:results_task_1}
\centering
\resizebox{\textwidth}{!}{
\begin{tabular}{l cc cc cc cc}
    \hline
    \\[-6pt]
    Model & \multicolumn{2}{c}{CPS Triton} & \multicolumn{2}{c}{CPS Poseidon} & \multicolumn{2}{c}{OSS Ceto} & \multicolumn{2}{c}{Average} \\
    \hline
    \\[-6pt]
    & MAE (\(\pm\) std) & R² (\(\pm\) std) & MAE (\(\pm\) std) & R² (\(\pm\) std) & MAE (\(\pm\) std) & R² (\(\pm\) std) & MAE (\(\pm\) std) & R² (\(\pm\) std) \\
    \hline
    \\[-6pt]
    
    Polynom & $.027 (\pm .001)$ & $.687 (\pm .015)$ & $.085 (\pm .010)$ & $.909 (\pm .037)$ & $.044 (\pm .007)$ & $.472 (\pm .465)$ & $.052 (\pm .006)$ & $.689 (\pm .172)$ \\
    CatBoost & $.019 (\pm .000)$ & $.840 (\pm .023)$ & $.067 (\pm .005)$ & $.935 (\pm .023)$ & $.043 (\pm .009)$ & $.506 (\pm .313)$ & $.043 (\pm .005)$ & $.760 (\pm .120)$ \\
    MLP & $.021 (\pm .001)$ & $.804 (\pm .040)$ & $.066 (\pm .002)$ & $.933 (\pm .025)$ & $.044 (\pm .005)$ & \bm{$.546 (\pm .277)$} & $.044 (\pm .003)$ & \bm{$.761 (\pm .114)$} \\
    TabPFN (500) & $.018 (\pm .000)$ & $.844 (\pm .029)$ & $.063 (\pm .003)$ & $.937 (\pm .027)$ & \bm{$.041 (\pm .005)$} & $.489 (\pm .393)$ & $.041 (\pm .003)$ & $.757 (\pm .150)$ \\
TabPFN (1000) & \bm{$.017 (\pm .000)$} & \bm{$.859 (\pm .022)$} & \bm{$.061 (\pm .002)$} & \bm{$.942 (\pm .019)$} & $.042 (\pm .004)$ & $.427 (\pm .513)$ & \bm{$.040 (\pm .002)$} & $.742 (\pm .184)$ \\

    \hline
    \\[-6pt]
\end{tabular}
}
\end{table*}

\paragraph{Task 2: Timeseries Regression} With temporal features included we observe in Table \ref{tab:results_task_2} that TabPFN again performed best on both cruise ships, achieving the lowest MAE and highest R². LSTM consistently ranked lowest, especially on the CPS Triton. On OSS Ceto, CatBoost slightly outperformed TabPFN in both MAE and R², while also showing the most stable results. TabPFN with 500 samples showed the weakest performance on this vessel. LSTM achieved a relatively high R² here, but with greater variability, indicating potential for sequence models with further optimization.
For OSS Ceto, CatBoost slightly outperformed TabPFN in terms of MAE (0.041 vs. 0.042), but also showed the best R² (0.558) with the lowest variance (±0.241). In contrast, TabPFN with 500 samples had the lowest R² (0.445) and the highest variance (±0.47), indicating reduced stability. The LSTM ranked lower in MAE but achieved a relatively high mean R² (0.537), though with higher variability.

\begin{table*}[ht]
\centering
\captionsetup{justification=centering}  % center the caption
\caption{Results of task 2 (Time-Series Regression): Comparison of the evaluated models in terms of MAE and R² metrics across the three vessel datasets. The values are averaged over the 5 cross-validation folds. The standard deviation is given in brackets. The error improves on some vessels compared to task 1. TabPFN slightly outperforms the other models in most cases.}
\label{tab:results_task_2}
\resizebox{\textwidth}{!}{
\begin{tabular}{l cc cc cc cc}
    \hline
    \\[-6pt]
    Model & \multicolumn{2}{c}{CPS Triton} & \multicolumn{2}{c}{CPS Poseidon} & \multicolumn{2}{c}{OSS Ceto} & \multicolumn{2}{c}{Average} \\
    \hline
    \\[-6pt]
    & MAE (\(\pm\) std) & R² (\(\pm\) std) & MAE (\(\pm\) std) & R² (\(\pm\) std) & MAE (\(\pm\) std) & R² (\(\pm\) std) & MAE (\(\pm\) std) & R² (\(\pm\) std) \\
    \hline
    \\[-6pt]
    
    CatBoost & $.017 (\pm .001)$ & $.867 (\pm .026)$ & $.064 (\pm .004)$ & $.940 (\pm .021)$ & \bm{$.041 (\pm .007)$} & \bm{$.557 (\pm .240)$} & $.041 (\pm .004)$ & \bm{$.788 (\pm .096)$} \\
    LSTM & $.020 (\pm .000)$ & $.812 (\pm .031)$ & $.072 (\pm .001)$ & $.927 (\pm .025)$ & $.045 (\pm .005)$ & $.536 (\pm .334)$ & $.046 (\pm .002)$ & $.758 (\pm .130)$ \\
    TabPFN (500) & $.017 (\pm .000)$ & $.869 (\pm .015)$ & $.062 (\pm .010)$ & $.940 (\pm .016)$ & $.042 (\pm .007)$ & $.445 (\pm .472)$ & $.040 (\pm .006)$ & $.751 (\pm .168)$ \\
    TabPFN (1000) & \bm{$.016 (\pm .001)$} & \bm{$.878 (\pm .025)$} & \bm{$.059 (\pm .005)$} & \bm{$.947 (\pm .015)$} & $.041 (\pm .008)$ & $.515 (\pm .326)$ & \bm{$.038 (\pm .004)$} & $.780 (\pm .122)$ \\

    \hline
    \\[-6pt]
\end{tabular}
}
\end{table*}

\paragraph{Summary of Key Findings} Across all tasks and vessel types, TabPFN consistently achieved the best performance, particularly when sufficient training data was available. Including weather and temporal information improved prediction accuracy, confirming the importance of contextual and sequential features. While model performance varied by vessel, the CPS Triton showed the lowest errors overall, and OSS Ceto posed greater modeling challenges due to operational complexity. These findings provide a clear basis for assessing future methods on this benchmark.

% Include content from sections/discussion.tex
\section{Discussion and Limitations} \label{discussion}

\paragraph{Temporal Context and Predictive Value}
The performance improvement from task 1 to task 2 suggests that incorporating temporal context provides additional predictive value. This is likely due to the model's ability to capture acceleration and deceleration phases, which span multiple observations and significantly affect fuel consumption. This effect is particularly pronounced in vessels such as CPS Triton and OSS Ceto, which operate in short, frequent voyages where transitional dynamics are more prominent.

\paragraph{Strong Performance of TabPFN Even With Limited Data}
TabPFN consistently performed well across all vessel types and on average, even with only 500-1000 training samples, highlighting its suitability for data-scarce time series scenarios. Its pretrained, inference-time-only architecture suggests that certain temporal patterns in fuel consumption can be captured without extensive task-specific training and can be computed directly on the ship. This supports the promise of foundation models for forecasting in domains like maritime transport, where labeled data is limited and real-world deployment often demands fast adaptability.

\paragraph{Cross-Vessel Generalization and Operational Variability}
Model performance was generally consistent across different vessel types, supporting the hypothesis that shared temporal structures in fuel consumption exist. However, notable deviations—especially for the OSS Ceto stemmed from a single cross-validation fold with distinct operational behavior. This underlines the need for careful dataset partitioning and suggests that model evaluation should account for behavioral regimes rather than just vessel categories.

\paragraph{Limitations}
Despite the promising results, several limitations should be acknowledged. The datasets, although varied, are relatively small in absolute terms and focused on two ship types, potentially limiting generalizability. We did not investigate predictions into the far future. For task 2 we limited our experiments to a fixed time context. Only a fixed set of model architectures and hyperparameters were explored. Additional tuning, especially for MLP and LSTM, might improve results. TabPFN was only trained on a subset of samples due to computational limitations and we only used simple sampling techniques to select the subset. Furthermore, only standard regression metrics were considered, without deeper evaluation of model uncertainty and explainability.

\paragraph{Applications}
Our framework has several promising applications in the context of data-driven analysis and decision-making on temporal ship operation data. It enables simulation of the ship model under arbitrary conditions to perform optimization of the fuel efficiency. In particular, this allows performing what-if szenarios with the ship to compare the consumption under different conditions. Our tests demonstrate that computation for both szenarios can be performed directly on the ship. In addition, analysts can use the framework to analyse consumption in the past and identify reasons for overconsumption. Based on the analysis they can optimize consumption and reduce emissions on future voyages.

\paragraph{Future Work}
Future work will focus on implementing k-step-ahead prediction using an autoregressive evaluation setup, particularly with exogenous inputs. K-step ahead prediction allows a ship operator to forecast fuel consumption over the next several time steps based on planned actions and expected conditions, enabling more informed decisions than current reactive approaches that rely solely on present or past data. We also plan to investigate additional model architectures such as Informer and TimesNet to evaluate potential improvements in handling complex temporal dynamics. This includes testing TabPFN with more samples and improved sample selection. To enhance generalizability, the approach will be extended to a wider range of vessel types and a larger fleet. Further improvements are expected through more effective hyperparameter tuning and a systematic study of input window size optimization.

% Include content from sections/conclusion.tex
\section{Conclusion} \label{conclusion}

In this work, we presented a new dataset of operational and environmental time-series data from three ships and introduced a novel benchmark covering tabular and time-series regression tasks.
Our results show that incorporating temporal context improves accuracy in some cases. TabPFN slightly outperformed other models, indicating the potential of in-context learning for ship fuel consumption prediction. In particular, in-context learning requires only little data to perform well. We also confirm that fuel consumption is influenced not only by vessel speed but also by environmental conditions such as weather and sea state.

Overall, our standardized benchmark provides a reproducible basis for evaluating temporal regression methods in the maritime domain and demonstrates the feasibility of modern machine learning - especially foundation models - for accurate onboard fuel estimation.

% Include content from sections/appendix.tex
%\input{sections/appendix}

\bibliography{ecai}
\end{document}